\documentclass[letterpaper]{article}
\usepackage{aaaifull}
\usepackage{times}
\usepackage{helvet}
\usepackage{courier}

\usepackage{mathrsfs}
\usepackage{mathptmx}
\usepackage{amsmath}
\usepackage{bbm}
\usepackage{longtable}
\usepackage{multirow}
\usepackage{array}
\usepackage{makecell}
\usepackage{amsfonts}

\newtheorem{theorem}{Theorem}[section]
\newtheorem{lemma}[theorem]{Lemma}
\newtheorem{proposition}[theorem]{Proposition}
\newtheorem{corollary}[theorem]{Corollary}
\newenvironment{proof}[1][Proof]{\begin{trivlist}

\newcommand{\qed}{\nobreak \ifvmode \relax \else
      \ifdim\lastskip<1.5em \hskip-\lastskip
      \hskip1.5em plus0em minus0.5em \fi \nobreak
      \vrule height0.75em width0.5em depth0.25em\fi}
\item[\hskip \labelsep {\bfseries #1}]}{\end{trivlist}}

\begin{document}
%
\setlength{\titlebox}{2.5in}

\title{Agent Behavior Prediction and Its Generalization Analysis}
\author{Fei Tian\thanks{This work was done when the first two authors were visiting Microsoft Research Asia.}\\
University of Science and Technology of China\\
tianfei@mail.ustc.edu.cn
\And
Haifang Li$^*$\\
Chinese Academy of Sciences\\
lihaifang@amss.ac.cn
\And
Wei Chen \\
Microsoft Research\\
wche@microsoft.com
\AND
Tao Qin\\
Microsoft Research\\
taoqin@microsoft.com
\And
Enhong Chen\\
University of Science and Technology of China\\
cheneh@ustc.edu.cn
\And
Tie-Yan Liu\\
Microsoft Research\\
tyliu@microsoft.com
}
\maketitle
\begin{abstract}
\begin{quote}
Machine learning algorithms have been applied to predict agent behaviors in real-world dynamic systems, such as advertiser behaviors in sponsored search and worker behaviors in crowdsourcing. The behavior data in these systems are generated by live agents: once the systems change due to the adoption of the prediction models learnt from the behavior data, agents will observe and respond to these changes by changing their own behaviors accordingly. As a result, the behavior data will evolve and will not be identically and independently distributed, posing great challenges to the theoretical analysis on the machine learning algorithms for behavior prediction. To tackle this challenge, in this paper, we propose to use Markov Chain in Random Environments (MCRE) to describe the behavior data, and perform generalization analysis of the machine learning algorithms on its basis. Since the one-step transition probability matrix of MCRE depends on both previous states and the random environment, conventional techniques for generalization analysis cannot be directly applied. To address this issue, we propose a novel technique that transforms the original MCRE into a higher-dimensional time-homogeneous Markov chain. The new Markov chain involves more variables but is more regular, and thus easier to deal with. We prove the convergence of the new Markov chain when time approaches infinity. Then we prove a generalization bound for the machine learning algorithms on the behavior data generated by the new Markov chain, which depends on both the Markovian parameters and the covering number of the function class compounded by the loss function for behavior prediction and the behavior prediction model. To the best of our knowledge, this is the first work that performs the generalization analysis on data generated by complex processes in real-world dynamic systems.
\end{quote}
\end{abstract}

\section{Introduction}

In this Internet era, more and more data are generated by self-interested agents in interactive systems. For example, in sponsored search, advertisers generate a large volume of bidding log data in their daily competitions with each other in attracting search users to click their ads; in crowdsourcing, workers generate a lot of behavior data when competing with other workers in getting tasks from the employers, and when completing the tasks assigned to them.

In many real cases including the aforementioned ones, there are three kinds of players in the systems: platform, users, and self-interested agents. Platform is the owner of the system, who designs the mechanism of the system and takes care of its execution. Users arrive at the platform in random, with their particular needs to be fulfilled. Agents behave strategically in order to attract the attention of the users so as to realize their own utilities. Taking both user needs and agent behaviors into consideration, the platform matches users with agents, extracts revenue from this procedure, and gives agents feedback on their performances (which depend on both the behaviors of agents and randomness in users). Upon the feedback, agents will adjust their behaviors in order to be better off in the future. To design a good mechanism, it is very important for the platform to understand and predict agent behaviors. With accurate prediction of agent behaviors, the platform can also provide tools to help agents to optimize their performances and therefore attract more agents to the system. Thus, predicting agent behaviors is an important task for the platform. For ease of reference, we call the problem of predicting agent behaviors in an interactive system \emph{``agent behavior prediction (ABP)''}.

\subsection{Examples of ABP}

Here we give three concrete examples of ABP for illustration, including systems of sponsored search, crowdsourcing and app stores.

In a sponsored search system, platform, users, and agents correspond to the search engine, search users, and advertisers respectively. Advertiser behaviors are the bid prices on their ads. When a search user issues a query, the search engine will run a GSP auction \cite{gsp} among all advertisers who bid on the query, rank their ads according to the product of the bid price and predicted click-through rate, and then charge the winning advertisers if the user clicks on their ads. After a period of time, the search engine will provide feedback to the advertisers about their performances (which we usually call Key Performance Indicators, or KPIs for short). The KPIs usually contain the numbers of impressions and clicks, the average rank positions, and the costs per click of their ads. Many advertisers will adjust their bid prices based on the feedback they receive, either by themselves or with the help of third-party search engine marketing companies. By logging the bid prices in a sufficiently long period of time, the search engine can predict how advertisers behave, and consequently enhance its click prediction algorithm and auction mechanism.

In crowdsourcing systems, platform, users, and agents correspond to crowdsourcing platform (e.g., \emph{Amazon Mechanical Turk}), employers, and workers respectively. A worker' behaviors usually include the claimed profile and expertise, the committed available time slots, the minimum payment requirement, and the quality of fulfilling a task. When an employer submits his/her task, the crowdsourcing platform will select workers according to some criteria, and then dispatch the tasks to them. After the workers finish the tasks, the employer will pay certain amount of money, part of which goes to the platform as commission and the rest goes to the workers. The employer will also rate the quality of the task completion. After a period of time, the platform will provide workers with necessary feedback, which usually includes the number of task assigned to them, the average ratings they receive, the total payments they obtain, etc. Some workers will adjust their behaviors based on the feedback in hope to be assigned more tasks and to get better rating/payment. With the logs of the behaviors of the workers, the platform can predict how workers behave, and enhance the worker engagement tool and task assignment mechanism on its basis.

In app stores, platform, users and agents correspond to app store, app users, and app developers respectively. The behaviors of the app developers include creating, describing, upgrading, and pricing their apps. When an app user searches for apps with certain functionalities, the app store will recommend a ranked list of related apps according to their claimed functionalities, popularities, reviews and ratings, prices, etc. The user might choose one or several apps from the list to install. For paid apps, the user needs to pay a certain amount of money, part of which goes to the app store as its commission, and the rest goes to the app developers.  After using these apps, the user may comment on the quality of the apps by submitting reviews and ratings. After a period of time, the app store will provide app developers with feedback, which usually includes the number of users who view their apps, the number of users who purchase and install their apps, the average rating and the number of reviews of their apps, and the total revenue they obtained from the installations of their apps. Many app developers will upgrade their apps (e.g., adjust the functionalities, descriptions, and prices) according to the feedback in hope to get a better performance in the future. With the logs of the behaviors of app developers, the app store can predict how developers behave and enhance its recommendation algorithm, and revenue sharing mechanism on its basis.

\subsection{Generalization Analysis for ABP}
Because of its importance, ABP has been studied in many works, including \cite{Greedy,PKey,Vindictive,xu2013predicting,ijcai/HeCWL13}. Some of them \cite{xu2013predicting,ijcai/HeCWL13} have adopted machine learning techniques and attempted to learn an agent behavior model by means of empirical risk minimization (ERM) on the behavior logs. Empirical results have shown that these machine learning techniques can significantly outperform previous non-learning approaches. However, despite the experimental success, it still remains an open question whether the use of ERM algorithms in behavior prediction is theoretically sound, and whether certain generalization ability of such algorithms can be guaranteed.

As far as we know, the answers to the above questions are unclear yet. This is mainly because of the complication of the corresponding theoretical analysis. As aforementioned, the behavior data are generated by self-interested agents, and dependent on both their previous behaviors and the random user factors in the system. As a result, the behavior data have quite complex statistical properties, and the generalization analysis in such a setting goes beyond the state of the art of statistical learning theory \cite{vapnik1998SLT,devroye1996probabilistic,yu1994rates}.

\subsection{Related Work}
\label{relatedwork}

There have been extensive game-theoretic models on advertisers' bidding prediction \cite{Greedy}\cite{Knapsack}\cite{Vindictive}. These models generally assume advertisers are fully rational and have full information access. This assumption is far beyond reality and thus recently machine learning methods \cite{BidForcasting}\cite{xu2013predicting}\cite{ijcai/HeCWL13}, based on minimizing prediction loss on advertisers' historical bidding data, have been employed in this task. For example, in \cite{xu2013predicting}, the authors propose models to describe different levels of advertisers' rationalities and fit the model parameters by learning from bidding data. The work most relevant to ours is \cite{ijcai/HeCWL13}, in which a Markov bidding model is introduced and a linear prediction function is learnt using Maximum Likelihood Estimation. However, whether these machine learning algorithms enjoy generalization ability remains an open theoretical problem since historical behavioral data are non i.i.d.

Previously machine learning researchers have established several theories for learning from non i.i.d data \cite{yu1994rates}\cite{vidyasagar2003learning}\cite{mohri2007stability}. In \cite{yu1994rates} and \cite{Zou:2009:GPE:1541660.1541688}, the authors establish
uniform convergence bounds for stationary process and
strongly mixing sequence. There are also theories specially built for learning under Markov chains such as \cite{Zou:2009:GPE:1541660.1541688}\cite{ContinumMarkov}, however they are not qualified to answer our question because they do not cover the case of learning under MCRE, a special Markov chain with time-variant transition probabilities. There are also works on MCRE in statistics literature such as \cite{MCRE0}\cite{MCRE1} and \cite{MCRE2}, but they mainly focus on statistical properties of MCRE such as ergodicity and central limit theorem. To the best of our knowledge there has been no previous works on generalization analysis of learning on data from MCRE.

\subsection{Our Results}

In order to analyze the ERM algorithms on agent behavior prediction, we propose a set of new techniques in this paper.

First, we model the generation process of the behavior data using a so-called \emph{Markov Chain in Random Environments} (MCRE), whose transition matrix is time-variant (depending on the random environments). Take sponsored search as an example. After the current-round auction, the advertiser observes his/her KPIs, which depend on both the bids of all the advertisers and the random clicks of the users. Based on the KPIs, the advertiser will determine how to set his/her own bid for the next round of auction and for different KPI values, the conditional distribution of his/her bid at the next round will be different. In this sense, the sequence of advertiser bids can be regarded as an MCRE.

Second, considering that it is difficult to perform generalization analysis on MCRE, we propose an equivalence transformation that maps the original MCRE to a higher dimensional time-homogenous Markov chain. Although the new Markov chain involves more variables, it is more \emph{regular} and thus easier to deal with from the perspective of generalization analysis. We prove that the new Markov chain will converge when time approaches infinity and a Hoeffding-style inequality holds for the empirical process associated with it.

Third, by exploiting the covering number technique, we derive a uniform convergence bound for the ERM algorithms on the data generated by the new Markov chain (and thus by the original MCRE due to the equivalence transformation). As a consequence, we prove that the ERM algorithms on the data generated by MCRE have their theoretical guarantees, which explains their good empirical performances reported in previous work. To the best of our knowledge, this is the first work in the literature that performs formal generalization analysis on the agent behavior prediction problem.

\section{Agent Behavior Prediction}
\label{model}
In this section we give a formal description of the Agent Behavior Prediction (ABP) problem. We first show that the generation process for the behaviors of self-interested agents can be described by a Markov Chain in Random Environments (MCRE), and then formulate ABP as an optimization problem.

\subsection{Agent Behaviors: Markov Chain in Random Environments}
The dynamic interactive systems mentioned in the introduction share some common properties. (1) The behaviors of an agent only depend on a finite number of his/her historical actions due to the limited memory of human being. In other words, the behaviors have Markovian properties. (2)Random users¡¯ behaviors are independent and identically distributed (i.i.d), for example, in sponsor search, there are two aspects associated with users: queries issued by users, and users¡¯ click patterns on ad ranking lists. It is clear that queries can be regarded as i.i.d. random variables. Click patterns are defined with respect to all possible ad ranking lists and they are also independent of agent behaviors(which only determine the selected ranking list), and can be regarded as i.i.d. (3) The behavior change of an agent is mainly affected by the feedback given by the platform. Since the feedback depends on the users randomly arriving at the system, the behavior change is not governed by a constant rule, but instead by some random factors.

Taking all the three properties into consideration, we can regard agent behaviors as generated by a Markov Chain in Random Environments (MCRE)\cite{MCRE0}. Note that this generation process is much more complicated than a simple i.i.d. sampling process or a time-homogeneous Markov process.

Before formally describing the MCRE process for agent behaviors, we make some assumptions. First, we assume that both the behavior space and feedback space are finite. This assumption is reasonable, since in many applications the behaviors of the agents are either characterized by categorical profiles (e.g., expertise and functionalities) or bounded and associated with a minimum unit (e.g., bidding price, payment requirements, and available time slots). The same reason holds for the finiteness of the feedback space, since feedback usually takes discrete values (e.g., number of clicks, ratings, and number of reviews) as well. Second, we assume there is a deterministic function that generates the feedback for a given agent $i$ based on the behaviors of all the agents and the random arrival of the users. This assumption is also reasonable because the feedback is usually provided by the platform using a deterministic algorithm.

With the above assumptions, now let us describe the generation process of the agent behavior according to \cite{ijcai/HeCWL13}. Suppose there are $N$ agents. Let $\mathbb{B}$ and $\mathbb{H}$ be the behavior space and feedback space of a single agent respectively, and $\mathbb{U}$ be the random factor space induced by users. We use the mapping $\eta_i: \mathbb{U}\times \mathbb{B}^N\rightarrow \mathbb{H}$ to denote the deterministic function that generates the feedback for agent $i$, and $\eta:\mathbb{U}\times \mathbb{B}^N\rightarrow \mathbb{H}^N$ to denote the joint feedback for all the agents. At the beginning of the $(t+1)$-th time period, agent $i$ receives feedback $h_t^i=\eta_i(u_t,b_t)$ about his/her behaviors in the $t$-th time period. Based on the feedback, agent $i$ may change his/her behavior to $b_{t+1}^i$ in order to be better off. That is, given the Markov property,

\begin{small}
\begin{equation}
P(b_{t+1}^{i}|b_1,...,b_t;u_1,...,u_t)=P(b_{t+1}^{i}|b_t^i;h_t^i).
\end{equation}
\end{small}

Note that the above equation implies the one-step Markov property, which is motivated here mainly due to ease of statement. Our analysis in this paper can be extended to higher order Markov chain as well, without too many modifications.

Given the feedback $h_t$, all the agents change their behaviors independently, so we have
\begin{small}
\begin{equation}
\label{MCRE}
P(b_{t+1}|b_1,...,b_t;u_1,...,u_t)=\prod_{i=1}^NP(b_{t+1}^i|b_t^i;h_t^i)=P(b_{t+1}|b_t,h_t).
\end{equation}
\end{small}This indicates that $\{b_t\}$ is a MCRE \cite{MCRE0}, where the environmental process is $\{h_t\}$. According to (\ref{MCRE}), the one-step transition matrix of an MCRE is time-variant and depends on the environmental variable.

\subsection{Learning Agent Behavior Model}

There exist some related works that leverage the empirical risk minimization (ERM) framework to learn the agent behavior model. Mathematically speaking, given a training set containing the behaviors of agents and the feedback they received in $T$ rounds, $\left\{\left(h_1,b_1\right);\left(h_2,b_2\right);...;\left(h_T,b_T\right)\right\}$, one aims to learn a function $f$: $\mathbb{H}^N\times \mathbb{B}^N\rightarrow \mathbb{B}^N$, which takes the behaviors and feedback at the current round as inputs and predicts the behavior at the next round. To this end, one minimizes the empirical risk on the training set: $\min_{f\in \mathcal{F}}\sum_{t=1}^Tl\left(f\left(h_t,b_t\right);b_{t+1}\right)$, where $l$ measures the loss between the predicted behavior and the real behavior in the training data. For example, $l$ can be the $0-1$ classification loss: $l\left(f\left(b_t,h_t\right);b_{t+1}\right)=\mathbbm{1}_{f\left(h_t,b_t\right)\neq b_{t+1}}$, and can also be some surrogate loss functions.

This ERM framework covers the algorithms for predicting advertiser behaviors in many previous works including \cite{ijcai/HeCWL13} and \cite{xu2013predicting}. For example, in \cite{ijcai/HeCWL13}, a truncated Gaussian function is used to model the Markov transition probabilities and a negative likelihood function is used as the loss function $l$. For another example, in \cite{xu2013predicting}, a compound function that considers the willingness, capability, and constraint of an advertiser is used as the model $f$, and again the negative likelihood of the historical behaviors is used as the loss function $f$.

As mentioned in the introduction, the ERM algorithms led to experimental success in the ABP tasks. However, it is unclear whether these algorithms have desired theoretical guarantee. In particular, it is unknown (1) qualitatively whether the ABP problem is learnable through an ERM process, and (2) quantitatively what is the relationship between generalization error and the size of the training data. The reason why the answers to these important questions are missing lies in that statistical learning theory mainly addresses learning problems with data generated by an i.i.d. sampling or a $\beta$-mixing Markov chain. This motivates us to formally study the learning theory with respect to the data generated by more complicated stochastic processes like MCRE.

\section{Generalization Bounds for ABP}
\label{SLT}

In this section, we perform generalization analysis on the ERM algorithms for agent behavior prediction. Our main result is stated in Theorem (\ref{uniformConvergence}). We prove the theorem in three steps: (1) constructing a new Markov chain of higher dimensionality but with more regular properties than the original MCRE; (2) proving the convergence of the empirical loss to the expected loss when the data are generated by this new Markov chain; (3) proving a uniform convergence bound by further leveraging the techniques of covering number. For ease of reference, we use a Notation Table \ref{tbl_symbol} in the end of the paper to summarize all new notations used in this section.

\subsection{Constructing a Higher-Dimensional Markov Chain}

The difficulty of analyzing the ERM algorithms when the data are generated by an MCRE lies in its time-variant transition probabilities. To tackle the challenge, we construct a higher-dimensional chain $\mathbb{M}=\left\{\left(h_t,b_t,b_{t+1}\right); t\ge 0\right\}$, by grouping \emph{correlated} variables together. Let $M_k(m,n)$ be one-step transition probability of $\{b_t\}$ from state $m$ to $n$ under the environment $k$. For convenience, we set $z_t=(h_t,b_t,b_{t+1})$ to be the $t$-th state of $\mathbb{M}$. Since the state space $\mathbb{H}^N\times\mathbb{B}^{2N}$ of the chain is finite, we label all the state values as $1,2...Z$, i.e., $\mathbb{M}$ takes value in state space $[Z]:=\{1,2,...,Z\}$.

The following Lemma (\ref{homogeneousLemma}) and Theorem (\ref{limExistTheorem}) show that the new Markov chain is time-homogeneous and has a stationary distribution under some mild assumptions.

\begin{lemma}
\label{homogeneousLemma}
We assume the random factors caused by users ${u_t}$ are i.i.d., then the constructed Markov chain $\mathbb{M}=\left\{\left(h_t,b_t,b_{t+1}\right); t\ge 0\right\}$ is time-homogeneous.

\begin{proof}
Since both \begin{itshape} feedback space\end{itshape} and \begin{itshape} behavior space
\end{itshape} are finite, the set of their three dimensional Cartesian product values is also finite.

To show that the process is a time-homogeneous Markov chain, we consider any two states $(j,p,q), (k,m,n) \in\mathbb{H}^N\times\mathbb{B}^{2N}$ and write their one-step transition probability as:

\begin{equation}
\label{TransMat}
\begin{aligned}
&P(h_t=k,b_t=m,b_{t+1}=n| \\
&(h_s,b_s,b_{s+1})_{s=1}^{t-2};\quad h_{t-1}=j,b_{t-1}=p,b_t=q)\\
=&P(h_t=k,b_t=m,b_{t+1}=n|h_{t-1}=j,b_{t-1}=p,b_t=q)\\
=&\left\{ \begin{array}{rcl}
 0 & \mbox{for} & m\neq q \\
 P(u\in \mathbb{U}:\eta(u,m)=k)M_k(m,n) & \mbox{for} & m=q
\end{array}\right.
\end{aligned}
\end{equation}
From (\ref{TransMat}), we know that the value of $(h_t,b_t,b_{t+1})$ only depends on $(h_{t-1},b_{t-1},b_{t})$. Therefore the process is a Markov chain. Furthermore, the one-step transition probability only depends on the states $(j,p,q), (k,m,n)$ and is independent of time index $t$, and therefore the transition probability is time-invariant.
\qed
\end{proof}
\end{lemma}

For ease of statement, we will use $M$ to denote the probability transition matrix in (\ref{TransMat}). As we can see from (\ref{TransMat}), lots of elements in $M$ are zero. Therefore it is not straightforward to judge whether Markov chain $\mathbb{M}$ will converge or not. Theorem (\ref{limExistTheorem}) shows that under some mild assumptions the convergence can be achieved.

\begin{theorem}
\label{limExistTheorem}
The Markov chain $\mathbb{M}$ composed by $\{z_t\}_{t=1}^T:=\{h_{t},b_{t},b_{t+1}\}_{t=1}^T$ has a stationary distribution under the following assumptions: \textbf{(A.1)} for every fixed value of the feedback $h_t=k$, the Markov process with transition matrix $M_k$ is irreducible and aperiodic; \textbf{(A.2)} the feedback function $\eta$ and the random user distribution satisfy $\forall m\in \mathbb{B}, k\in \mathbb{H}, P(u\in \mathbb{U}:\eta(u,m)=k)>0$. \footnote{Condition (A.2), which seems not very intuitive, basically says that every possible value of the feedback is \emph{reachable} from every value of the behavior, if there are a very large number of random users arriving at the platform and they have very high dynamics and variety.}
\end{theorem}

To prove this theorem, since the state of $(h_{t},b_t,b_{t+1})$ is finite and $\mathbb{M}$ is time-homogeneous, we only need to prove that Markov chain $\mathbb{M}$ is irreducible and aperiodic, which is shown in the following two lemmas respectively.

\begin{lemma}
Under the assumption $(A.1)$ and $(A.2)$, the Markov chain $\mathbb{M}$ is irreducible.
\begin{proof}

According to the definition of irreducibility, we need to show that any two states $(k,m,n)\in \mathbb{H}^N\times\mathbb{B}^{2N}$ and $(j,p,q)\in \mathbb{H}^N\times\mathbb{B}^{2N}$ are accessible to each other. To prove that, we show three simple facts in the following:

\begin{itemize}

\item According to assumption $(A.2)$, it is possible to produce feedback signals $k$ by a one step transition from state $(j,p,q)$. i.e. $\exists x\in \mathbb{B}, s.t. P(z_{t+1}=(k,q,x)|z_t=(j,p,q))>0$.

\item In Markov chain $M_k$ where the feedback signal is fixed to be $k$, there exists $d\in\{1,2,\cdots\}$, such that we can build a $d$-step transition path from behavior profile $x$ to behavior profile $m$, followed by a one step transition to behavior profile $n$.

The existence of the $d$-step path from $q$ to $m$ is due to the irreducibility of $M_k$ in assumption $(A.1)$. To see that it is possible to transit from state $m$ to $n$ by one step in Markov chain $M_k$, note that if $M_{k}(m,n)=0$, we can simply erase the state $(k,m,n)$ from the state space of Markov chain $\mathbb{M}$, which does not affect our results, therefore we only need to consider the case in which $M_{k}(m,n)>0$.

\item  According to assumtion $(A.2)$, in Marokov chain $\mathbb{M}$, it is possible to observe $d+1$ consecutive states in which the $d+1$ feedback signals are all $k$.

Combining the three points, we can obtain:

\begin{equation}
\begin{aligned}
&P(z_{t+d+2}=(k,m,n)|z_t=(j,p,q))\\
\geq & P(z_{t+d+2}=(k,m,n)|z_{t+1}=(k,q,x)) \\
&\times P(z_{t+1}=(k,q,x)|z_t=(j,p,q))\\
=&M_k(m,n)\cdot P(u\in \mathbb{U}:\eta(u,m)=k) \\
&\times \cdots \times M_k(q,x)\cdot P(u\in \mathbb{U}:\eta(u,q)=k)\\
>& 0.
\end{aligned}
\end{equation}

\end{itemize}

Therefore state $(k,m,n)$ is reachable from state $(j,p,q)$. Similarly, we can also prove that state $(j,p,q)$ is reachable from state $(k,m,n)$. Since $(k,m,n)$ and $(j,p,q)$ are arbitrarily chosen, we actually prove the irreducibility of the Markov chain $\mathbb{M}$.
\qed
\end{proof}
\end{lemma}

\begin{lemma}
Under the assumption $(A.1)$ and $(A.2)$, the Markov chain $\mathbb{M}$ is aperiodic.
\begin{proof}
Since the Markov chain is irreducible, all states in the chain have the same period. Therefore, in order to prove the aperiodicity, we just need to show that for a given state $(k,m,m)\in \mathbb{H}^N\times\mathbb{B}^{2N}$, its period is one. According to the first assumption in the lemma, $\forall d\geq1$ satisfying $M_k^{(d)}(m,m)>0$. Therefore we can build a $d$-step path in Markov chain $M_k$: $m\rightarrow m_1\rightarrow m_2\cdots \rightarrow
m_{d-1}\rightarrow m$, such that the transition probability in every step is positive, i.e., $M_k(m,m_1)>0,\ M_k(m_1,m_2)>0,\ ...,\ M_k(m_{d-1},m)>0$. As a result,

\begin{equation}
\begin{aligned}
&P(z_{t+d}=(k,m,m)|z_t=(k,m,m))\\
\geq &P(z_{t+d}=(k,m,m)|z_{t+d-1}=(k,m_{d-1},m)) \\
 &\times P(z_{t+d-1}=(k,m_{d-1},m)|z_{t+d-2}=(k,m_{d-2},m_{d-1})) \\
&\times \cdots \times P(z_{t+1}=(k,m,m_1)|z_t=(k,m,m))\\
=&M_k(m_{d-1},m)\cdot P(u\in \mathbb{U}:\eta(u,m_{d-1})=k) \\
&\times M_k(m_{d-2},m_{d-1})\cdot P(u\in \mathbb{U}:\eta(u,m_{d-2})=k) \\
&\times \cdots \times M_k(m,m_1)\cdot P(u\in \mathbb{U}:\eta(u,m)=k)\\
>& 0.
\end{aligned}
\end{equation}

Hence, a $d$-step transition path with positive probability in Markov chain $\mathbb{M}$ can be built as $(k,m,m)\rightarrow (k,m,m_1)\rightarrow (k,m_1,m_2)\cdots \rightarrow (k,m_{d-1},m)\rightarrow (k,m,m)$. Then by the definition of \emph{period} \footnote{In Markov chain $P$, state $i$'s period is defined as the g.c.d. of all $d\in{1,2,\cdots}$ satisfying $P^{(d)}(i,i)>0$.}, Markov chain $\mathbb{M}$ has the same period as Markov chain $M_k$, which is one.\qed
\end{proof}
\end{lemma}

\subsection{Convergence Bound}

In the previous subsection, we have constructed a new Markov chain and proved its convergence. In this subsection, we show that these results can be used to analyze the convergence rate of the empirical risk for a specified behavior prediction model, namely $f$.

Let us start from a formal definition of the problem. Given the $T$-round training data $S=\{(h_1,b_1,b_2);(h_2,b_2,b_3);\cdots;(h_T,b_T,b_{T+1})\}=(z_1,z_2,\cdots,z_T)$, we define the $T$-round empirical risk of $f$ with respect to $S$ as follows:

\begin{equation}
\label{defineOfEmpiricalLoss}
err_S^T(f)=\frac{1}{T}\sum_{t=1}^Tl(f(b_t,h_t),b_{t+1})=\frac{1}{T}\sum_{t=1}^Tl(f,z_t),
\end{equation}
where we assume $l$ to be upper bounded by a constant $B>0$.

Since the Markov chain $\{z_t\}$ is ergodic, we assume that it is a stationary Markov chain with unique stationary distribution $\pi$. Then the expected risk of $f$ is
\begin{equation}
\label{defineOfExpectedLoss}
err_{\pi}(f)= E_{\pi}l(f,z)=\frac{1}{T}\sum_{t=1}^TE_{\pi}l(f,z_t).
\end{equation}
We refer to $err_{\pi}(f)$ as the \emph{real} expected risk of $f$.

Next we investigate how well the $T$-round empirical risk $err_S^T(f)$ approximates $err_{\pi}(f)$. For this purpose, we leverage the Hoeffding Inequality for uniformly ergodic Markov Chains \cite{glynn2002hoeffding}, which is rephrased as below for completeness.

\begin{proposition} (\textbf{Hoeffding Inequality for uniformly ergodic Markov Chains})
\label{hoeffProp}
Let $X=(X_n:n\geq 0)$ be a Markov Chain taking values in a state space $S$, if the following assumption holds: \textbf{(A.3)} there exists a probability measure $\phi$ on $S$, $\lambda >0$, and an integer $m\geq 1$ s.t. $\forall x\in S$, $P(X_m\in \cdot|X_0=x)\geq \lambda\phi(\cdot)$, then for a function $g:S\rightarrow \mathcal{R}$ with its norm defined as $||g||=sup\{|g(x)|: x\in S\}<\infty$, define $S_T=\frac{1}{T}\sum_{t=1}^Tg(X_t)$, for $T>2||g||m/(\lambda\epsilon)$, we have
\begin{small}
\begin{equation}\label{hoeffForMar}
P\left(\left|S_T-E(S_T)\right|\geq \epsilon\right)\leq 2exp(-\frac{\lambda^2(T\epsilon-2||g||m/\lambda)^2}{2T||g||^2m^2}),
\end{equation}
\end{small}where the expectation $E(S_T)$ is taken on the stationary distribution of $X$.
\end{proposition}

In order to leverage Proposition (\ref{hoeffProp}), we need to check whether its assumption $(A.3)$ holds in our problem. For this purpose, we note the fact that for an irreducible, aperiodic, and finite-state Markov chain with time-invariant transition probability matrix $P$, there must exist $N$ such that $\forall n\geq N$, all elements of $n$-step transition matrix $P^{(n)}$ are non-zero( Lemma 6.6.3 in \cite{durrett2010probability} ). Accordingly in our setting, for Markov chain $\mathbb{M}$, there exists $N_0$ such that $\forall 1\leq i,j\leq Z, M^{(N_0)}_{i,j}> 0$. Denote $\delta$ as the minimum element in $M^{(N_0)}$, i.e., $\delta=\min\limits_{1\leq i,j\leq Z}M^{(N_0)}_{i,j}>0$. Then if we set $m=N_0$, $\lambda=Z\delta$ and set $\phi$ to be the uniform distribution on $[Z]$, it is easy to see that $(A.3)$ holds. Further noticing that $||g||$ in (\ref{hoeffForMar}) is $B$ in our setting, where $g=l(f;z_t)$ and $B$ is the upperbound of function $l$, we can leverage Proposition (\ref{hoeffProp}) to obtain desired convergence bound as Theorem (\ref{ConvergenceTheorem}) shows.

\begin{theorem} \textbf{Convergence of Empirical Loss to Expected Loss}
\label{ConvergenceTheorem}
Let $f:\mathbb{H}^N\times \mathbb{B}^N\rightarrow \mathbb{B}^N$ be the behavior prediction function, $err_S^T(f)$ and $err_{\pi}(f)$ be the empirical loss and expected loss respectively, as defined in equation (\ref{defineOfEmpiricalLoss}) and (\ref{defineOfExpectedLoss}). For any $\epsilon>0$ and $T>2BN_0/(Z\delta\epsilon)$, we have
\begin{small}
\begin{equation}
\begin{aligned}
P(\left|err_S^T(f)-err_{\pi}(f)\right|\geq\epsilon)\leq 2exp\left(-\frac{Z^2\delta^2\left(T\epsilon-2BN_0/(Z\delta)\right)^2}{2TB^2N_0^2}\right)
\end{aligned}
\end{equation}
\end{small}
\end{theorem}Theorem (\ref{ConvergenceTheorem}) basically states that when the sample size $T$ is large enough, the empirical risk $err_S^T(f)$ will converge to the long-term expected risk $err_{\pi}(f)$, and the convergence rate is exponential in the sample size.

\subsection{Uniform Convergence Bound}

Considering that uniform ergodic Markov chains are $\beta$-mixing \cite{doob1953stochastic}, in this subsection, we prove a uniform convergence bound for the ABP problem based on the \emph{independent block technique} for $\beta$-mixing sequences \cite{Yu94:yu1994rates}  and covering number technique\footnote{Covering number is one of the common ways to measure the complexity of a function class. Specifically, covering number $\mathcal{N}(\epsilon,l\circ\mathcal{F},T)$ is defined as $\max\limits_{\{z_t\}_{t=1}^T\in [Z]^T}\mathcal{N}(\epsilon,l\circ\mathcal{F}|_{\{z_t\}_{t=1}^T},d_T)$ in which $\mathcal{N}(\epsilon,l\circ\mathcal{F}|_{\{z_t\}_{t=1}^T},d_T)$ is the minimum capacity of $\epsilon$-cover of $(l\circ\mathcal{F})$'s projection on data ${\{z_t\}_{t=1}^T}$, w.r.t. the distance metric between $x\in R^T, y\in R^T$ defined as $d_T(x,y):=\max\limits_{1\leq t\leq T}\left|x_t-y_t\right|$.}   The independent block technique transforms the original problem based on dependent samples to that based on independent blocks. Then, we are able to apply the symmetrization technique and Hoeffding inequality to obtain the desirable bound as shown in the following theorem.

\begin{theorem}\textbf{Uniform Convergence Theorem.}
\label{uniformConvergence}
Let $\mathcal{F}=\{f:\mathbb{H}^N\times \mathbb{B}^N\rightarrow \mathbb{B}^N\}$ be the behavior function space and $l\circ\mathcal{F}$ be the composite function set of the loss function $l$ acting on $\mathcal{F}$. For a behavior function $f\in\mathcal{F}$, denote $err_S^T(f)$ and $err_{\pi}(f)$ as its empirical loss and expected loss respectively, as defined in equation (\ref{defineOfEmpiricalLoss}) and (\ref{defineOfExpectedLoss}). For any $\epsilon>0$, $m \in \mathbb{N}_+$, $\tau \in \mathbb{N}_+$, with $2\tau m = T$, we have
\begin{small}
\begin{equation}
\begin{aligned}
&P\left(\sup\limits_{f\in\mathcal{F}}|err_S^T(f)-err_{\pi}(f)|\geq\epsilon\right)\\
\leq &16\mathcal{N}(\epsilon/16, l \circ \mathcal{F},T)\exp(- \frac{\epsilon^2\tau}{128B^2})+2\tau \beta(z,m)
\end{aligned}
\end{equation}
\end{small}where $B$ is the upper bound of loss function $l(f,z)$, and $\beta(z,m)$ is the $\beta$-mixing rate of $\{z_t\}$.
\end{theorem}

The proof of the theorem contains two steps. For the first step, we employ the independent block technique to transform the original problem based on dependent samples to that based on independent blocks. For the second step, we further apply the symmetrization technique and Hoeffding inequality to obtain the desirable bound.
\begin{proof}
Divide $\{z_t\}$ into $2 \tau$ blocks, each of which consists of $m$ consecutive samples. For $1 \leq j \leq \tau$, we define the time index sets $H_j = \{t: 2(j-1)m + 1 \leq t \leq (2j-1)m\}$ and $T_j = \{t: (2j-1)m + 1 \leq t \leq (2j)m\}$. We introduce i.i.d. blocks $\{\tilde{z}_t: t \in H_j\}$, and they have the same distribution with $\{z_t: t \in H_1\}$. Then we have,
\begin{equation}
\label{eqn_1}
\begin{aligned}
&P\left(\sup\limits_{f\in\mathcal{F}}|err_S^T(f)-err_{\pi}(f)|\geq\epsilon\right)\\
\leq &2P\left(\sup\limits_{f\in\mathcal{F}}|\frac{1}{\tau}\sum_{j=1}^{\tau}\left(\sum_{t \in H_j} l(f, z_t) - E \sum_{t \in
H_j}l(f, z_t)\right)| \geq m\epsilon\right)\\
\leq &2P\left(\sup\limits_{f\in\mathcal{F}}|\frac{1}{\tau}\sum_{j=1}^{\tau}\left(\sum_{t \in H_j} l(f, \tilde{z}_t) - E \sum_{t \in
H_j}l(f, \tilde{z}_t)\right)| \geq m\epsilon\right) + 2 \tau \beta(z,m)\\
\end{aligned}
\end{equation}
Since $\{\sum_{t \in H_j} l(f, \tilde{z}_t\}_{j=1}^{\tau}$ are i.i.d., we can apply the symmetrization technique by introducing i.i.d. ghost samples $\{\hat{z}_t: t \in H_j\}$, we have
\begin{equation}
\label{eqn_2}
\begin{aligned}
&P\left(\sup\limits_{f\in\mathcal{F}}|\frac{1}{\tau}\sum_{j=1}^{\tau}\left(\sum_{t \in H_j} l(f, \tilde{z}_t) - E \sum_{t \in
H_j}l(f, \tilde{z}_t)\right)| \geq m\epsilon\right)\\
\leq &2P\left(\sup\limits_{f\in\mathcal{F}}|\frac{1}{\tau}\sum_{j=1}^{\tau}\sum_{t \in H_j} l(f, \tilde{z}_t) - \frac{1}{\tau}\sum_{j=1}^{\tau}\sum_{t \in H_j} l(f, \hat{z}_t)| \geq m\epsilon/2\right)
\end{aligned}
\end{equation}
Now we introduce Rademacher variables $\{\sigma_j\}_{j=1}^{\tau}$, which are i.i.d., and satisfying $P(\sigma_j = \pm 1)= 1/2, j=1,...,\tau.$ Then we have
\begin{equation}
\label{eqn_3}
\begin{aligned}
&P\left(\sup\limits_{f\in\mathcal{F}}|\frac{1}{\tau}\sum_{j=1}^{\tau}\sum_{t \in H_j} l(f, \tilde{z}_t) - \frac{1}{\tau}\sum_{j=1}^{\tau}\sum_{t \in H_j} l(f, \hat{z}_t)| \geq m\epsilon/2\right)\\
=&P\left(\sup\limits_{f\in\mathcal{F}}|\frac{1}{\tau}\sum_{j=1}^{\tau}\sigma_j(\sum_{t \in H_j} l(f, \tilde{z}_t) - \sum_{t \in H_j} l(f, \hat{z}_t))| \geq m\epsilon/2\right)\\
\leq&2P\left(\sup\limits_{f\in\mathcal{F}}|\frac{1}{\tau}\sum_{j=1}^{\tau}\sigma_j(\sum_{t \in H_j} l(f, \tilde{z}_t))| \geq m\epsilon/4\right)
\end{aligned}
\end{equation}
We use the covering number technique and hoeffding inequality to obtain our desirable results. Pick a subset $\mathcal{G}\subseteq\mathcal{F}$ such that $l\circ\mathcal{G}$ is an $\epsilon/16$-cover of $l\circ\mathcal{F}$ with respect to $l_{\infty}$ metric $d_{T}(l\circ f_1,l\circ f_2):=\max\limits_{1\leq t\leq T}\left|l(f_1,z_t)-l(f_2,z_t)\right|$. Then we have
\begin{equation}
\label{eqn_4}
\begin{aligned}
&P\left(\sup\limits_{f\in\mathcal{F}}|\frac{1}{\tau}\sum_{j=1}^{\tau}\sigma_j(\sum_{t \in H_j} l(f, \tilde{z}_t)) | \geq m\epsilon/4\right)\\
\leq & 2\mathcal{N}(\epsilon/16, l \circ \mathcal{F},T)\exp(- \frac{\epsilon^2\tau}{128B^2})
\end{aligned}
\end{equation}
Therefore by combining these inequalities (\ref{eqn_1}), (\ref{eqn_2}), (\ref{eqn_3}) and (\ref{eqn_4}), we obtain
\begin{equation}
\begin{aligned}
&P\left(\sup\limits_{f\in\mathcal{F}}|err_S^T(f)-err_{\pi}(f)|\geq\epsilon\right)\\
\leq &16\mathcal{N}(\epsilon/16,l \circ \mathcal{F},T)\exp(- \frac{\epsilon^2\tau}{128B^2})+2\tau \beta(z,m)
\end{aligned}
\end{equation}
\qed
\end{proof}

\textbf{Remark}: Please note that for most regular function class, the covering number $\mathcal{N}(\epsilon,l\circ\mathcal{F},T)$ defined in Theorem \ref{uniformConvergence} can be polynomially bounded. For example,
\begin{itemize}
\item If the loss function $l$ is Lipschitz-continuous in its first argument with Lipschitz constant $L>0$, then for any $T$, we have $\mathcal{N}(\epsilon,l\circ\mathcal{F},T)\leq\mathcal{N}(\epsilon/L,\mathcal{F},T)$.
\item Since we assume the behavior set to be finite, the ABP problem can actually be regarded as a multi-class classification problem, where the class number is $|\mathbb{B}|$. In this case, the covering number $\mathcal{N}(\frac{\epsilon}{8}/L,\mathcal{F},2T)$ can be bounded by the growth function of $\mathcal{F}$, defined as $\max\limits_{\{z_t\}_{t=1}^T\in [Z]^{2T}}\big|\mathcal{F}|_{\{z_t\}_{t=1}^{2T}}\big|$. Moreover, the growth function is bounded by $(\frac{2Te(|\mathbb{B}|+1)^2}{2d})^d$ \cite{SLT4MultiClass}, where $d$ is the Natarajan dimension of $\mathcal{F}$ \cite{NatarajanDim}. Thus $\mathcal{N}(\frac{\epsilon}{8}/L,\mathcal{F},2T)$ is at most in $T$'s polynomial order.
\end{itemize}

One may note that the value of $m$ affects the bound stated in Theorem (\ref{uniformConvergence}). The following corollary shows how to set the value of $m$ when $\{z_t\}$ forms an algebraically mixing process.

\begin{corollary}
If $\{z_t\}$ forms an algebraically mixing sequence, i.e., $\beta(z, m) \leq \beta_0 m^{- \gamma}$, where $\beta_0, \gamma\geq 0$, then the optimal value of $m$ is given by $m = C(T^{\frac{1}{1+s}})$, where $0 < s < \gamma$. For any $\epsilon > 0$, the uniform convergence bound becomes:
\begin{equation}
\begin{aligned}
&P\left(\sup\limits_{f\in\mathcal{F}}|err_S^T(f)-err_{\pi}(f)|\geq\epsilon\right)\\
\leq &16\mathcal{N}(\epsilon/16, l \circ \mathcal{F},T)\exp(- \frac{\epsilon^2}{256B^2C}T^{\frac{s}{1+s}})+ \beta_0 C^{-(r+1)}T^{\frac{s-\gamma}{1+s}}
\end{aligned}
\end{equation}
\end{corollary}

\section{Conclusion and Future Work}
In this paper, we have studied the generalization ability of ERM algorithms for agent behavior prediction. In particular, we first develop a new technique that transforms MCRE to a higher-dimensional but more regular Markov chain and then give a uniform generalization bound based on the new Markov chain. As for the future work, we plan to investigate the joint learning problem of the optimal mechanism of the platform and the optimal prediction model of agent behaviors. Generalization analyses for these two cases will be even more challenging, and the corresponding results will also have more profound impact on adopting machine learning techniques in real-world interactive systems.
\label{conclusion}
\label{notations}

\begin{table}[htpb]
\footnotesize
\centering
\begin{tabular}{|c|p{150pt}|}
\hline
Notation & \makecell[c]{Meaning} \\ \hline
$\mathbb{U},\mathbb{B},\mathbb{H}$ &Respectively the random user factors space, agent behavior space and feedback space \\ \hline
$u_t, b_t,h_t$ &Respectively the random users, agents joint behavior and feedback at round $t$ \\ \hline
$N$& Number of agents in the system\\ \hline
$\eta$ &Feedback function outputting feedbacks to all agents. $\eta$ maps from $\mathbb{U}\times\mathbb{B}^N$ to $\mathbb{H}^N$\\ \hline
$M_k$ & Agents joint behavior transition probability matrix under joint feedback $k$\\ \hline
$\mathbb{M}$ & The new construocted higher dimensional chain, $\mathbb{M}=\left\{\left(h_t,b_t,b_{t+1}\right); t\ge 0\right\}$\\ \hline
$Z$ &$\mathbb{M}$'s state number. We assume that $M$ takes states from $[Z]=\{1,2,...,Z\}$\\ \hline
$z_t$ &$\mathbb{M}'$'s state value at round $t$, $z_t=\{h_t,b_t,b_{t+1}\}$\\ \hline
$M$&Transition probability matrix of $\mathbb{M}$\\ \hline
$\pi$ &$\mathbb{M}$'s stationary distribution\\ \hline
$N_0$& The elements of $N_0$-step transition probability matrix $M^{(N_0)}$ of $\mathbb{M}$ are all positive\\ \hline
$\delta$ & Minimum element of $M^{(N_0)}$\\ \hline
$\mathcal{F}$& Function class of behavior prediction functions, $\mathcal{F}\subset\{f:\mathbb{H}^N\times\mathbb{B}^N\rightarrow \mathbb{B}^N\}$ \\ \hline
$l$ & Loss function w.r.t. behavior prediction function $f$ and data, e.g, $l(f,z_t)=\mathbbm{1}_{f(h_t,b_t)\neq b_{t+1}}$\\ \hline
$B$& Upper bound for loss function $l$\\ \hline
$S$, $\tilde{S}$ & $T$-round training data and ghost data sampled from $\mathbb{M}$ respectively, $S=(z_t)_{t=1}^T$, $\tilde{S}=(\tilde{z}_t)_{t=1}^T$\\ \hline
$err_S^T(f)$, $err_{\pi}(f)$&Respectively denotes the $T$-round empirical loss on $S$ and expected loss of an agent behavior prediction function $f$\\ \hline
\end{tabular}
\caption{Notations}
\label{tbl_symbol}
\end{table}

\section{Acknowledgement}
This work is partially supported by grant from the National Science Foundation for Distinguished Young Scholars of China (Grant No. 61325010). We thank Di He for his valuable suggestions on the detailed proof technique. Thanks to the AAAI anonymous reviewers for their comments to make the paper more exact and clear.
\newpage
\bibliography{ref}

\begin{thebibliography}{}

\bibitem[\protect\citeauthoryear{Bendavid \bgroup et al\mbox.\egroup
  }{1995}]{SLT4MultiClass}
Bendavid, S.; Cesabianchi, N.; Haussler, D.; and Long, P.~M.
\newblock 1995.
\newblock Characterizations of learnability for classes of (0,..., n)-valued
  functions.
\newblock {\em Journal of Computer and System Sciences} 50(1):74--86.

\bibitem[\protect\citeauthoryear{Cary \bgroup et al\mbox.\egroup
  }{2007}]{Greedy}
Cary, M.; Das, A.; Edelman, B.; Giotis, I.; Heimerl, K.; Karlin, A.~R.;
  Mathieu, C.; and Schwarz., M.
\newblock 2007.
\newblock Greedy bidding strategies for keyword auctions.
\newblock In {\em EC '07 Proceedings of the 8th ACM conference on Electronic
  commerce.}
\newblock ACM Press.

\bibitem[\protect\citeauthoryear{Chakrabarty, Zhou, and
  Lukose.}{2007}]{Knapsack}
Chakrabarty, D.; Zhou, Y.; and Lukose., R.
\newblock 2007.
\newblock Budget constrained bidding in keyword auctions and online knapsack
  problems.
\newblock In {\em WWW '07 Proceedings of the 16th international conference on
  World Wide Web.}
\newblock ACM Press.

\bibitem[\protect\citeauthoryear{Cogburn}{1980}]{MCRE0}
Cogburn, R.
\newblock 1980.
\newblock Markov chains in random environments: the case of markovian
  environments.
\newblock {\em The Annals of Probability}  908--916.

\bibitem[\protect\citeauthoryear{Cogburn}{1984}]{MCRE1}
Cogburn, R.
\newblock 1984.
\newblock The ergodic theory of markov chains in random environments.
\newblock {\em Zeitschrift f{\"u}r Wahrscheinlichkeitstheorie und verwandte
  Gebiete} 66(1):109--128.

\bibitem[\protect\citeauthoryear{Cogburn}{1991}]{MCRE2}
Cogburn, R.
\newblock 1991.
\newblock On the central limit theorem for markov chains in random
  environments.
\newblock {\em The Annals of Probability} 19(2):587--604.

\bibitem[\protect\citeauthoryear{Cui \bgroup et al\mbox.\egroup
  }{2011}]{BidForcasting}
Cui, Y.; Zhang, R.; Li, W.; and Mao., J.
\newblock 2011.
\newblock Bid landscape forecasting in online ad exchange marketplace.
\newblock In {\em KDD '11 Proceedings of the 17th ACM SIGKDD international
  conference on Knowledge discovery and data mining.}
\newblock ACM Press.

\bibitem[\protect\citeauthoryear{Devroye}{1996}]{devroye1996probabilistic}
Devroye, L.
\newblock 1996.
\newblock {\em A probabilistic theory of pattern recognition}, volume~31.
\newblock springer.

\bibitem[\protect\citeauthoryear{Doob}{1953}]{doob1953stochastic}
Doob, J.
\newblock 1953.
\newblock {\em Stochastic Processes}.
\newblock Wiley Publications in Statistics. John Wiley \& Sons.

\bibitem[\protect\citeauthoryear{Durrett}{2010}]{durrett2010probability}
Durrett, R.
\newblock 2010.
\newblock {\em Probability: theory and examples}, volume~3.
\newblock Cambridge university press.

\bibitem[\protect\citeauthoryear{Edelman, Ostrovsky, and Schwarz}{2005}]{gsp}
Edelman, B.; Ostrovsky, M.; and Schwarz, M.
\newblock 2005.
\newblock Internet advertising and the generalized second price auction:
  Selling billions of dollars worth of keywords.
\newblock Technical report, National Bureau of Economic Research.

\bibitem[\protect\citeauthoryear{Glynn and Ormoneit}{2002}]{glynn2002hoeffding}
Glynn, P.~W., and Ormoneit, D.
\newblock 2002.
\newblock Hoeffding's inequality for uniformly ergodic markov chains.
\newblock {\em Statistics \& probability letters} 56(2):143--146.

\bibitem[\protect\citeauthoryear{He \bgroup et al\mbox.\egroup
  }{2013}]{ijcai/HeCWL13}
He, D.; Chen, W.; Wang, L.; and Liu, T.-Y.
\newblock 2013.
\newblock A game-theoretic machine learning approach for revenue maximization
  in sponsored search.

\bibitem[\protect\citeauthoryear{Mohri and
  Rostamizadeh}{2007}]{mohri2007stability}
Mohri, M., and Rostamizadeh, A.
\newblock 2007.
\newblock Stability bounds for non-iid processes.
\newblock In {\em Advances in Neural Information Processing Systems},
  1025--1032.

\bibitem[\protect\citeauthoryear{Natarajan}{1989}]{NatarajanDim}
Natarajan, B.~K.
\newblock 1989.
\newblock On learning sets and functions.
\newblock {\em Machine Learning} 4(1):67--97.

\bibitem[\protect\citeauthoryear{Pin and Key.}{2011}]{PKey}
Pin, F., and Key., P.
\newblock 2011.
\newblock Stochasitic variability in sponsored search auctions: Observations
  and models.
\newblock In {\em EC '11 Proceedings of the 12th ACM conference on Electronic
  commerce.}
\newblock ACM Press.

\bibitem[\protect\citeauthoryear{Vapnik}{1998}]{vapnik1998SLT}
Vapnik, V.~N.
\newblock 1998.
\newblock Statistical learning theory.

\bibitem[\protect\citeauthoryear{Vidyasagar}{2003}]{vidyasagar2003learning}
Vidyasagar, M.
\newblock 2003.
\newblock {\em Learning and generalization: with applications to neural
  networks}.
\newblock Springer.

\bibitem[\protect\citeauthoryear{Xu \bgroup et al\mbox.\egroup
  }{2013}]{xu2013predicting}
Xu, H.; Gao, B.; Yang, D.; and Liu, T.-Y.
\newblock 2013.
\newblock Predicting advertiser bidding behaviors in sponsored search by
  rationality modeling.
\newblock In {\em Proceedings of the 22nd international conference on World
  Wide Web},  1433--1444.
\newblock International World Wide Web Conferences Steering Committee.

\bibitem[\protect\citeauthoryear{Yu}{1994a}]{yu1994rates}
Yu, B.
\newblock 1994a.
\newblock Rates of convergence for empirical processes of stationary mixing
  sequences.
\newblock {\em The Annals of Probability}  94--116.

\bibitem[\protect\citeauthoryear{Yu}{1994b}]{Yu94:yu1994rates}
Yu, B.
\newblock 1994b.
\newblock Rates of convergence for empirical processes of stationary mixing
  sequences.
\newblock {\em The Annals of Probability}  94--116.

\bibitem[\protect\citeauthoryear{Zhang and Tao}{2012}]{ContinumMarkov}
Zhang, C., and Tao, D.
\newblock 2012.
\newblock Generalization bounds of erm-based learning processes for
  continuous-time markov chains.
\newblock {\em IEEE Trans. Neural Netw. Learning Syst.}  1872--1883.

\bibitem[\protect\citeauthoryear{Zhou and Lukose.}{2007}]{Vindictive}
Zhou, Y., and Lukose., R.
\newblock 2007.
\newblock Vindictive bidding in keyword auctions.
\newblock In {\em ICEC '07 Proceedings of the ninth international conference on
  Electronic commerce.}
\newblock ACM Press.

\bibitem[\protect\citeauthoryear{Zou, Li, and
  Xu}{2009}]{Zou:2009:GPE:1541660.1541688}
Zou, B.; Li, L.; and Xu, Z.
\newblock 2009.
\newblock The generalization performance of erm algorithm with strongly mixing
  observations.
\newblock {\em Mach. Learn.} 75(3):275--295.

\end{thebibliography}
\bibliographystyle{aaai}

\clearpage

\end{document}